\documentclass[letterpaper, 10 pt, conference]{ieeeconf}      
\usepackage{graphics}
\usepackage{graphicx}
\usepackage{bm}
\usepackage{indentfirst}
\usepackage{multirow}
\usepackage{cite}
\usepackage{url}
\usepackage[ruled,vlined]{algorithm2e}
\usepackage{makecell}
\usepackage{amsmath,amssymb,amsfonts}
\usepackage{booktabs}
\usepackage[margin=2cm]{geometry}
\usepackage{xcolor}
\usepackage{caption}
\usepackage[compatible]{algpseudocode} 
\usepackage{threeparttable}
\usepackage{float}

\IEEEoverridecommandlockouts                              

\title{\LARGE \bf
Effective Reinforcement Learning Control using\\ Conservative Soft Actor-Critic
}

\author{Zhiwei Shang$^{1 \dag}$, Xinyi Yuan$^{2 \dag}$, Wenjun Huang$^{3}$,  Yunduan Cui$^{3}$, Di Chen$^{4}$, Meixin Zhu$^{5*}$
\thanks{\dag Zhiwei Shang and Xinyi Yuan have contributed equally to this work.}
\thanks{$^*$ Corresponding author: Meixin Zhu (e-mail: meixin@ust.hk)}
\thanks{$^{1}$ School of Data Science, The Chinese University of Hong Kong, Shenzhen, China.}
\thanks{$^{2}$ Graduate School of Engineering Science, Osaka University, Japan.}
\thanks{$^{3}$ Shenzhen Institute of Advanced Technology, Chinese Academy of Sciences, China.}
\thanks{$^{4}$ Department of Electrical and Electronic Engineering, The Hong Kong Polytechnic University, Hong Kong SAR, China.}
\thanks{$^{5}$ School of Transportation, Southeast University, China.}
}

\begin{document}

\maketitle
\thispagestyle{empty}
\pagestyle{empty}

\begin{abstract}
Reinforcement Learning (RL) has shown great potential in complex control tasks, particularly when combined with deep neural networks within the Actor-Critic (AC) framework. However, in practical applications, balancing exploration, learning stability, and sample efficiency remains a significant challenge. Traditional methods such as Soft Actor-Critic (SAC) and Proximal Policy Optimization (PPO) address these issues by incorporating entropy or relative entropy regularization, but often face problems of instability and low sample efficiency.
In this paper, we propose the Conservative Soft Actor-Critic (CSAC) algorithm, which seamlessly integrates entropy and relative entropy regularization within the AC framework. CSAC improves exploration through entropy regularization while avoiding overly aggressive policy updates with the use of relative entropy regularization. Evaluations on benchmark tasks and real-world robotic simulations demonstrate that CSAC offers significant improvements in stability and efficiency over existing methods. These findings suggest that CSAC provides strong robustness and application potential in control tasks under dynamic environments.
\end{abstract}

\section{Introduction}\label{S1}

Reinforcement Learning (RL) has emerged as a powerful method for learning optimal control strategies in complex and dynamic environments without relying on prior knowledge~\cite{sutton2018reinforcement}, with broad success in robotic control and game playing~\cite{kober2013reinforcement, mnih2015human, silver2018general, kaufmann2023champion, wang2024arm, miao2024effective}. When deep neural networks~\cite{goodfellow2016deep} are combined with the Actor-Critic (AC) architecture~\cite{lillicrap2016continuous}, RL can effectively handle high-dimensional state and action spaces, achieving superhuman performance across challenging domains~\cite{zhu2020dynamic, cui2021autonomous, chen2023risk, yan2021hybrid, cai2024deep, zhou2024t}. However, deploying RL in real-world systems requires careful balancing of sample efficiency, learning stability, and control performance~\cite{Zhu2020The, dulac2021challenges, kroemer2021review}. To address this, researchers have explored entropy and relative entropy regularization to optimize the exploration-exploitation trade-off.

Soft Actor-Critic (SAC)~\cite{haarnoja2018soft}, a representative Maximum Entropy RL method, uses entropy regularization to encourage exploration and has achieved strong results in various control tasks~\cite{10098549,wahid2021learning,he2022multiagent,nakhleh2023sacplanner}. However, maximizing policy entropy can destabilize learning, leading to overly aggressive updates when policy-entropy dynamics are misaligned or the environment changes significantly. In contrast, Trust Region Policy Optimization (TRPO)~\cite{schulman2015trust} and Proximal Policy Optimization (PPO)~\cite{schulman2017proximal} use relative entropy constraints to limit update magnitudes, improving stability but often at the cost of sample efficiency in on-policy settings~\cite{xue2023proximal, sadhukhan2021multi, guan2020centralized, bohn2019deep}.

\begin {table}[t]
\caption{Comparison of CSAC and the related approaches} \label{table:1}
\begin{center}
    \begin{tabular}{ | l | c | c | }
    \hline
    Approach & Entropy & Relative Entropy\\ \hline
    TD3~\cite{fujimoto2018addressing} & $\times$ & $\times$  \\\hline
    PPO~\cite{schulman2017proximal} & $\times$ & $\bigcirc$ \\\hline
    SAC~\cite{haarnoja2018soft} & $\bigcirc$ & $\times$ \\\hline
    CSAC (ours) & $\bigcirc$ & $\bigcirc$ \\\hline
    \end{tabular}
\end{center}
\vspace{-12mm}
\end{table}

Despite these advances, few works have integrated both entropy and relative entropy regularization within an AC framework for continuous control. Prior methods such as Dynamic Policy Programming (DPP)~\cite{azar2012dynamic} and Conservative Value Iteration (CVI)~\cite{kozuno2019theoretical} have demonstrated the benefits of combining these regularization techniques in value-based settings, with extensions further validating their potential~\cite{cui2017kernel, tsurumine2019deep, shang2021shiftable, shang2023relative}. However, the computational complexity of SoftMax operations in continuous action spaces has largely confined these methods to discrete settings.

To address this gap, we propose Conservative Soft Actor-Critic (CSAC), which integrates entropy and relative entropy regularization within the AC framework. CSAC uses entropy regularization in the critic's Q-function to promote exploration, while relative entropy regularization constrains policy updates to ensure stability and robustness. As summarized in Table~\ref{table:1}, the contributions of this paper are:
\begin{enumerate}
\item We develop an RL method that combines entropy and relative entropy regularization, unifying the stability of PPO with the exploration capability of SAC, and demonstrate significant improvements in sample efficiency and learning stability across various control tasks.
\item We extend the principles of CVI and DPP to continuous action spaces within the AC framework without introducing additional computational complexity, broadening the applicability of entropy-constrained RL to complex continuous control tasks.
\item Through extensive experiments on benchmark tasks and Real-Robot-Based simulations, we validate that CSAC significantly outperforms existing RL algorithms in learning stability, sample efficiency, and final policy performance.
\end{enumerate}

\section{Preliminaries}\label{S2}
\subsection{Markov Decision Processes}\label{S2-1}
The interaction between the environment and RL agent is formulated as a Markov Decision Process (MDP) with a 5-tuple $(\mathcal{S}, \mathcal{A}, r, \mathcal{P}, \gamma)$, where $\mathcal{S}$ and $\mathcal{A}$ denote the state and action spaces, $r(\bm{s}, \bm{a})$ is the reward function, $\mathcal{P}: \mathcal{S}\times\mathcal{S} \times \mathcal{A} \mapsto \left[0, 1\right]$ is the transition probability, and $\gamma \in [0, 1)$ is the discount factor. The policy $\pi:\mathcal{S} \times\mathcal{A}\mapsto [0, 1]$ maps states to a distribution over actions. The objective is to find an optimal policy maximizing the value function:
\begin{align}
V_{\pi}(\bm{s}) = \mathbb{E}_{\pi}\biggl[\sum_{t=0}^{\infty }\gamma ^tr(\bm{s}_t, \bm{a}_t)\biggr].
\label{eq: objective}
\end{align}
Let $\bm{s}^\prime\sim\mathcal{P}(\bm{s}^\prime|\bm{s}, \bm{a})$ and $\bm{a}^\prime\sim\pi(\bm{a}^\prime|\bm{s}^\prime)$. The corresponding state-action value function satisfies the Bellman equation:
\begin{align}
Q_{\pi}\left(\bm{s}, \bm{a}\right)=r\left(\bm{s}, \bm{a}\right)+\gamma \mathbb{E}_{\bm{s}^\prime, \bm{a}^\prime}\left[Q_{\pi}\left(\bm{s}^\prime, \bm{a}^\prime\right)\right].
\label{eq: Q_B}
\end{align}

\subsection{Actor-Critic Framework}\label{S2-2}

The Actor-Critic (AC) framework integrates policy-based and value-based approaches through two components: the Actor $\pi(\bm{a}|\bm{s}; \bm{\phi})$, which represents the policy, and the Critic $Q(\bm{s,a}; \bm{\theta})$, which estimates the action-value function. The Actor is updated by following the policy gradient using feedback from the Critic, while the Critic is trained to minimize the Temporal Difference (TD) error. This framework serves as the foundation for several advanced RL algorithms such as DDPG~\cite{lillicrap2015continuous}, SAC, and PPO, demonstrating its versatility in various control tasks.

\section{Approach}\label{S3}

\subsection{Mix Entropy and Relative Entropy in Value Function}\label{S3-1}
CSAC incorporates both entropy and relative entropy regularization into the value function. The augmented reward function is defined as:

\vspace{-2mm}
\begin{align}
r_{\text{CSAC}}(\bm{s}_t, \bm{a}_t) =& r(\bm{s}_t, \bm{a}_t) + \gamma\sigma \mathbb{E}_{\bm{s}_{t+1}} \left[ - \log (\pi(\cdot |\bm{s}_{t+1})) \right] \nonumber \\
& - \gamma\tau \mathbb{E}_{\bm{s}_{t+1}} \left[ \log \frac{\pi(\cdot | \bm{s}_{t+1})}{\widetilde{\pi}(\cdot | \bm{s}_{t+1})} \right],
\label{eq: r_CSAC}
\end{align}

where $\sigma$ and $\tau$ modulate the entropy and relative entropy terms, respectively. Substituting Eq.~\eqref{eq: r_CSAC} into the Bellman equation (Eq.~\eqref{eq: Q_B}) yields the CSAC value function:
\vspace{-2mm}
\begin{align}
Q_{\text{CSAC}}&\left(\bm{s}_t, \bm{a}_t\right)=r\left(\bm{s}_t, \bm{a}_t\right)+\gamma \mathbb{E}_{\bm{s}_{t+1},\atop \bm{a}_{t+1}}[Q_{\text{CSAC}}\left(\bm{s}_{t+1},\bm{a}_{t+1}\right)\nonumber\\& -\sigma \log\left(\pi (\bm{a}_{t+1} | \bm{s}_{t+1})\right)-\tau
\log \frac{\pi(\bm{a}_{t+1} | \bm{s}_{t+1})}{\widetilde{\pi}(\bm{a}_{t+1} | \bm{s}_{t+1})}]
\label{eq: Q_CSAC}
\end{align}

Following TD3~\cite{fujimoto2018addressing} and SAC~\cite{haarnoja2018soft}, we employ two critic networks $Q_{\text{CSAC}}^{\bm{\theta}_1}$ and $Q_{\text{CSAC}}^{\bm{\theta}_2}$ to mitigate overestimation. Given a sample $\{\bm{s}, \bm{a}, r(\bm{s}, \bm{a}), \bm{s}^{\prime}\}\sim\mathcal{D}$ with $\bm{a}^{\prime}\sim\pi(\cdot|\bm{s}^{\prime})$, the critic loss is:
\vspace{-2mm}
\begin{align}
L\left(\bm{\theta}_i\right)=\left[Q_{\text{CSAC}}^{\bm{\theta}_i}(\bm{s, a})-y(\bm{s}, \bm{a}, \bm{s}^{\prime})\right]^2, \quad i \in\{1,2\}
\label{eq: Q_TD_1}
\end{align}

where the TD target takes the minimum over two target networks (parameters $\hat{\bm{\theta}}$):
\vspace{-2mm}
\begin{align}
y\left(\bm{s}, \bm{a}, \bm{s}^{\prime}\right)=r(\bm{s}, \bm{a})+\gamma\left[\min_{j=1,2} Q_{\text{CSAC}}^{\hat{\bm{\theta}}_{j}}\left(\bm{s}^{\prime}, {\bm{a}}^{\prime}\right)\right.\nonumber\\\left.-\sigma \log (\pi ({\bm{a}}^{\prime} | \bm{s}^{\prime}))-\tau
\log \frac{\pi({\bm{a}}^{\prime} | \bm{s}^{\prime})}{\widetilde{\pi}({\bm{a}}^{\prime} |\bm{s}^{\prime})}\right]
\label{eq: Q_TD_2}
\end{align}

Here the previous policy $\widetilde{\pi}$ is iteratively stored and updated during learning to stabilize training.

\subsection{Policy Improvement with Mixed Entropy and Relative Entropy Regularization}\label{S3-2}
We now derive the actor update rule. Following SAC~\cite{haarnoja2018soft}, the optimal policy for any state $\bm{s}$ is:
\vspace{-2mm}
\begin{align}
\pi^* =& \arg \max _\pi \sum_{\bm{a} \in \mathcal{A}} \pi(\bm{a} | \bm{s})\bigg[Q_{\text{CSAC}}^*\left(\bm{s}, \bm{a}\right) \nonumber\\
& - \sigma \log (\pi (\bm{a} | \bm{s})) - \tau \log \frac{\pi(\bm{a} | \bm{s})}{\widetilde{\pi}(\bm{a} | \bm{s})}\bigg].
\label{eq: Pi_CSAC}
\end{align}

For continuous action spaces, we solve Eq.~\eqref{eq: Pi_CSAC} subject to $\int \pi(\bm{a}|\bm{s})d\bm{a} = 1$ via the Lagrangian:
\vspace{-2mm}
\begin{align}
\mathcal{L}(\bm{s},\lambda) &= \int\pi(\bm{a}|\bm{s})\bigg[Q_{\text{CSAC}}(\bm{s},\bm{a}) + \sigma\big[- \log (\pi(\bm{a} | \bm{s}))\big] \nonumber\\
& - \tau\big[\log \frac{\pi( \bm{a} | \bm{s} )}{\widetilde{\pi}(\bm{s} | \bm{s})}\big]\bigg]d\bm{a} - \lambda\left(\int\pi(\bm{a}|\bm{s})d\bm{a}-1\right).
\label{eq: Lag}
\end{align}

Setting $\partial \mathcal{L}/\partial \pi = 0$ and following CVI~\cite{kozuno2019theoretical} with $\alpha=\frac{\tau}{\tau+\sigma}$ and $\beta=\frac{1}{\tau+\sigma}$, we obtain:
\vspace{-2mm}
\begin{align}
\pi(\bm{a}|\bm{s}) = \widetilde{\pi}^{\alpha}(\bm{a}|\bm{s})e^{(-\beta\lambda-1)}e^{\beta Q_{\text{CSAC}}(\bm{s},\bm{a})}.
\label{eq: Lag_3}
\end{align}

Applying the normalization condition to solve for $\lambda$ and combining with Eq.~\eqref{eq: Lag_3}, the updated policy at iteration $k$ is:

\begin{align}
\pi_{k+1}(\bm{a} | \bm{s})=\frac{\pi_k^\alpha(\bm{a} | \bm{s}) \exp \left(\beta Q_{\text{CSAC}, k+1}(\bm{s}, \bm{a}) \right)}{\int\pi_k^\alpha(\bm{a} | \bm{s}) \exp \left(\beta Q_{\text{CSAC}, k+1}(\bm{s}, \bm{a})\right)d\bm{a}}.
\label{eq: Lag_policy}
\end{align}

Introducing the action preference function $P$~\cite{sutton2018reinforcement,azar2012dynamic}, this can be expressed as:

\begin{align}
P_{k+1}(\bm{s, a})&=\frac{\alpha}{\beta} \log \pi_k(\bm{a} | \bm{s}) + Q_{\text{CSAC}, k+1}(\bm{s},\bm{a}), \\
V_{\text{CSAC}, k}(\bm{s}) &= \frac{1}{\beta} \log \int \exp \left(\beta P_{k}(\bm{s, a})\right)d\bm{a}, \\
\pi_{k}(\bm{a} | \bm{s}) &= \exp\left(\beta\left(P_k\left(\bm{s,a}\right)-V_{\text{CSAC}, k}\left(\bm{s}\right)\right)\right).
\label{eq: Lag_policy_2}
\end{align}

Eq.~\eqref{eq: Lag_policy} conforms to the Energy-Based Policy (EBP) model~\cite{haarnoja2018soft}, which cannot be sampled directly. Following~\cite{haarnoja2017reinforcement2, haarnoja2018soft}, we approximate it with a Gaussian policy by minimizing the KL divergence. For $\bm{s}\sim\mathcal{D}$, the actor loss is:

\begin{align}
J_\pi(\bm{\phi}) &\!= \!D_{\mathrm{KL}}[\pi\left(\cdot \mid \bm{s}\right) \| \exp \beta(P^{\text{}}(\bm{s},\cdot)\!-\!V_{\text{CSAC}}^{\text{}}(\bm{s}))]\nonumber\\
&\! =\!\mathbb{E}_{\bm{a} \sim \pi}\!\left[\log \pi\left(\bm{a} | \bm{s}\right)\!-\!\beta P\left(\bm{s}, \bm{a}\right) \!+\!\beta V_{\text{CSAC}}(\bm{s})\right].
\label{eq: actor_update_1}
\end{align}

The actor generates actions via the reparameterization trick:

\begin{align}
\tilde{\bm{a}}(\bm{s}, \bm{\xi})=\tanh \left(\mu(\bm{s})+\sigma(\bm{s}) \odot \bm{\xi}\right), \quad \bm{\xi} \sim \mathcal{N}(\bm{0}, \bm{I}),
\label{eq: actor}
\end{align}

where $\mu(\bm{s})$ and $\sigma(\bm{s})$ are the mean and variance. Dividing Eq.~\eqref{eq: actor_update_1} by $\beta$ and omitting the intractable normalization term, the practical actor loss becomes:

\begin{align}
J_\pi({\bm{\phi}})&\approx(\tau +\sigma) \log \pi\left(\tilde{\bm{a}}(\bm{s}, \bm{\xi}) | \bm{s}\right) - \tau \log \widetilde{\pi}(\tilde{\bm{a}}(\bm{s}, \bm{\xi}) | \bm{s})\nonumber\\
& - \min_{i=1,2} (Q_{\text{CSAC}}^{\bm{\theta}_{i}}(\bm{s},\tilde{\bm{a}}(\bm{s}, \bm{\xi}))),
\label{eq: actor_update_2}
\end{align}

where the action preference function uses the minimum critic output:

\begin{align}
P\left(\bm{s}, \tilde{\bm{a}}(\bm{s}, \bm{\xi})\right) \! = \!\tau \!\log \widetilde{\pi}(\tilde{\bm{a}}(\bm{s}, \bm{\xi}) | \bm{s}) \!+\! \min_{i=1,2} (Q_{\text{CSAC}}^{\bm{\theta}_{i}}(\bm{s},\tilde{\bm{a}}(\bm{s}, \bm{\xi}))).
\label{eq: actor_update_3}
\end{align}

\begin{algorithm}[t]
\caption{Conservative Soft Actor-Critic}\label{alg:1}
Initialize memory buffer $\mathcal{D}$, soft update rate $\rho$, $\sigma$ and $\tau$ that control entropy and relative entropy terms, critic networks $Q^{\bm{\theta}_1}, Q^{\bm{\theta}_2}$, target critic networks  $Q^{\hat{\bm{\theta}}_1}, Q^{\hat{\bm{\theta}}_2}$ with copied parameters $\hat{\bm{\theta}}_1 \leftarrow \bm{\theta}_1, \hat{\bm{\theta}}_2 \leftarrow \bm{\theta}_2$, actor network $\pi_{\bm{\phi}}$ and previous actor networks $\pi^{\bm{\phi}_p}$ with current weights: $\bm{\phi}_p \leftarrow \bm{\phi}$\\
\For{$t=1$ to $T_{\text{initial}}$}{
\# Collect sample with a random policy\\
$\mathcal{D} \leftarrow\left(\bm{s}_{t}, \bm{a}_{t}, r(\bm{s}_t, \bm{a}_t), \bm{s}_{t+1}\right)$}

\For {$t = 1$ to $T$}{
\# Interaction phase\\
Observe state $\bm{s}_t$\\
Decide and execute action ${\bm{a}}_t \sim \pi^{\bm{\phi}}\left({\bm{a}}_t | {\bm{s}}_t\right)$\\
Observe next state $\bm{s}_{t+1}$ and get reward $r(\bm{s}_t, \bm{a}_t)$\\
$\mathcal{D} \leftarrow\left(\bm{s}_{t}, \bm{a}_{t}, r(\bm{s}_t, \bm{a}_t), \bm{s}_{t+1}\right)$\\
\# Update phase\\
$\bm{\phi}_p \leftarrow \bm{\phi}$\\
Random sample mini-batch of $M$ tuples from $\mathcal{D}$\\
\For{\text{each sampled tuple} $(\bm{s}, \bm{a}, r(\bm{s}, \bm{a}), \bm{s}^\prime)$}{
\# Calculate TD error following Eq.~\eqref{eq: Q_TD_2}\\
${\bm{a}}^{\prime} \sim \pi^{\bm{\phi}}\left(\cdot | \bm{s}^{\prime}\right)$\\
$\begin{aligned}&y(\bm{s}, \bm{a},\bm{s}^{\prime}) = r(\bm{s}, \bm{a})+\gamma(\min _{j=1,2} Q^{\hat{\bm{\theta}}_j}\left(\bm{s}^{\prime}, \bm{a}^{\prime}\right) \\ & -\sigma \log \left(\pi^{\bm{\phi}}\left(\bm{a}^{\prime} | \bm{s}^{\prime}\right)\right)-\tau \log \frac{\pi^{\bm{\phi}}\left(\bm{a}^{\prime} | \bm{s}^{\prime}\right)}{\pi_{\bm{\phi}_p}\left(\bm{a}^{\prime} | \bm{s}^{\prime}\right)})\end{aligned}$\\
\# Update critic networks following Eq.~\eqref{eq: Q_TD_1}\\
$\nabla_{\bm{\theta}_i}[Q^{\bm{\theta}_i}(\bm{s, a})-y(\bm{s}, \bm{a},\bm{s}^{\prime})]^2$, for $i \in\{1,2\}$\\
\# Update actor networks following Eq.~\eqref{eq: actor_update_2}\\
$\begin{aligned}&\nabla_{\bm{\phi}}(\tau +\sigma) \log \pi^{\bm{\phi}} (\tilde{\bm{a}}(\bm{s}, \bm{\xi}) | \bm{s})  - \\ &
\tau \log {\pi}^{\bm{\phi}_p}(\tilde{\bm{a}}(\bm{s}, \bm{\xi})_t | \bm{s})-\min _{i=1,2} (Q^{\bm{\theta}_{i}}(\bm{s},\tilde{\bm{a}}(\bm{s}, \bm{\xi})))\end{aligned}$\\
\# Update target networks weights\\
$\hat{\bm{\theta}}_i \leftarrow \rho \bm{\theta_i}+(1-\rho)\hat{\bm{\theta}}_i$ for $i \in\{1,2\}$}
}
\Return{$Q^{\bm{\theta}_1}(\cdot), Q^{\bm{\theta}_2}(\cdot), \pi^{\bm{\phi}}(\cdot)$}
\end{algorithm}

\vspace{-3mm}
\subsection{Summary of Conservative Soft Actor-Critic}\label{S3-3}

The complete CSAC algorithm is summarized in Algorithm~\ref{alg:1}. After initializing the networks and collecting initial samples with a random policy, CSAC alternates between an interaction phase (collecting experience into replay buffer $\mathcal{D}$) and an update phase. In each update, the previous actor parameters are stored, and a mini-batch is sampled from $\mathcal{D}$ to update the critic (Eq.~\eqref{eq: Q_TD_1}) and actor (Eq.~\eqref{eq: actor_update_2}) networks via gradient descent, followed by soft updates of the target networks controlled by $\rho$.

\section{Experiments}\label{S4}
\subsection{Experimental Settings}\label{S4-1}

\begin{figure}[H]
	\centering
	\includegraphics[width=1.0\columnwidth]{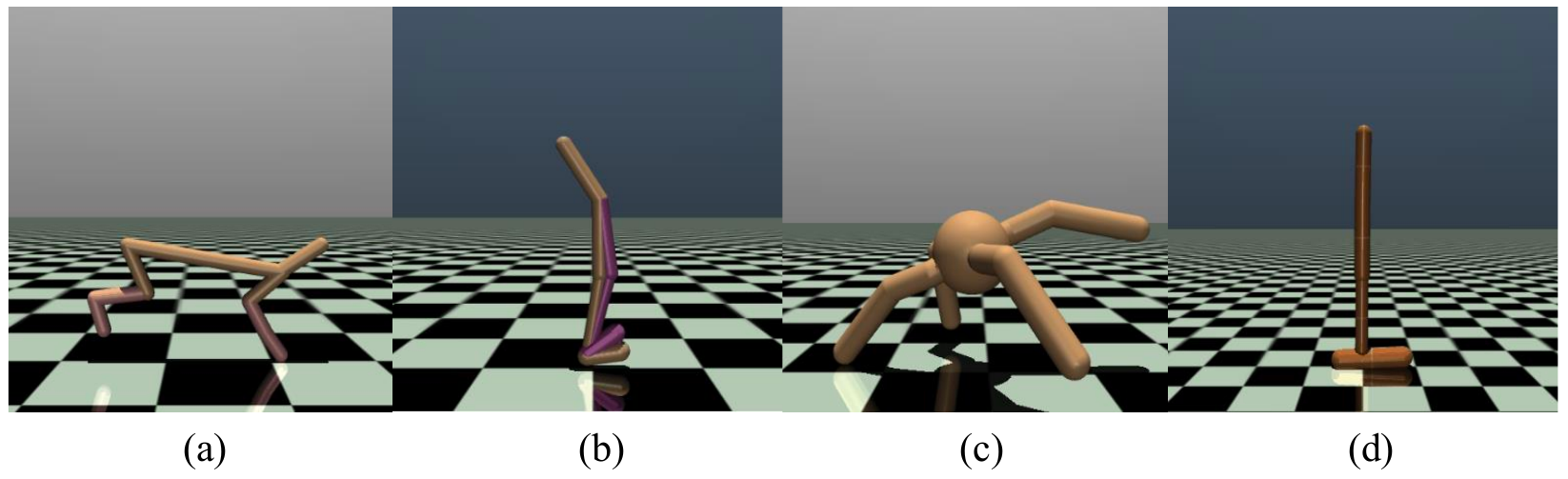}
	\caption{Benchmark tasks for evaluation in this article. (a) HalfCheetah-v4. (b) Walker2d-v4. (c) Ant-v4. (d) Hopper-v4. }
	\label{fig:mujoco}
\end{figure}

We evaluated CSAC on four MuJoCo benchmark tasks\cite{todorov2012mujoco, towers2024gymnasium}, including HalfCheetah-v4, Walker2d-v4, Ant-v4, and Hopper-v4, which cover diverse robotic structures and dynamic control challenges. The task specifications are shown in Fig.~\ref{fig:mujoco} and Table~\ref{t:mujoco}.

\begin{table}[H]
    \caption{Observation and Action Spaces for Different MuJoCo Tasks}
    \centering
    \label{table:spaces}
    \resizebox{1.0\linewidth}{!}{\begin{tabular}{|c|c|c|}
        \hline
        \textbf{} & \textbf{Observation Space} & \textbf{Action Space} \\\hline
        HalfCheetah-v4 & 17 & 6 \\\hline
        Walker2d-v4    & 17 & 6 \\\hline
        Ant-v4         & 27 & 8 \\\hline
        Hopper-v4      & 11 & 3 \\\hline
    \end{tabular}}
    \label{t:mujoco}
\end{table}

We compared CSAC against four baselines: SAC\cite{haarnoja2018soft}, PPO\cite{schulman2017proximal}, TD3\cite{fujimoto2018addressing}, and the recent SD3\cite{pan2020softmax}, spanning both on-policy and off-policy approaches. All results are averaged over five independent trials. CSAC was implemented using PaddlePaddle~\cite{ma2019paddlepaddle} with its RL toolkit PARL, and experiments were conducted on a server with an Intel Xeon Gold-6248R CPU, NVIDIA RTX-A5000 GPU, and 256~GB RAM (Ubuntu 20.04).

Common hyperparameters are listed in Table~\ref{table:parameters 1}. For CSAC, the entropy coefficient $\sigma$ was set to $0.2$ and the relative entropy coefficient $\tau$ to $0.5$ across all MuJoCo tasks.

\begin{table*}[t]
    \caption{COMMON HYPER-PARAMETER SETTINGS OF COMPARED APPROACHES}
    \centering
    \label{table:parameters}
    \resizebox{0.75\linewidth}{!}{\begin{tabular}{|c|c|c|c|c|c|}
        \hline
        & \textbf{CSAC(ours)} & \textbf{SAC} & \textbf{PPO} & \textbf{TD3} & \textbf{SD3} \\\hline
        Critic Learning Rate & $3\cdot10^{-4}$ & $3\cdot10^{-4}$ & $3\cdot10^{-4}$ & $3\cdot10^{-4}$ & $1\cdot10^{-3}$ \\\hline
        Actor Learning Rate & $3\cdot10^{-4}$ & $3\cdot10^{-4}$ & $3\cdot10^{-4}$ & $3\cdot10^{-4}$ & $1\cdot10^{-3}$ \\\hline
        Actor and Critic Structure & (256,256) & (256,256) & (64,64) & (256,256) & (256,256) \\\hline
        Target Update Rate  & 0.005 & 0.005 & / & 0.005 & 0.005 \\\hline
        Batch Size & 256 & 256 & 32 & 256 & 256 \\\hline
        Memory Size & $1\cdot10^{6}$ & $1\cdot10^{6}$ & / & $1\cdot10^{6}$ & $1\cdot10^{6}$ \\\hline
        Warmup Steps & $1\cdot10^{4}$ & $1\cdot10^{4}$ & / & $1\cdot10^{4}$ & $1\cdot10^{4}$ \\\hline
        Discount Factor  & 0.99 & 0.99 & 0.99 & 0.99 & 0.99 \\\hline
    \end{tabular}}
    \label{table:parameters 1}
\end{table*}

\subsection{Evaluation of Learning Capability on MuJoCo Tasks}\label{S4-2}

\begin{figure*}[t]
	\centering
	\includegraphics[width=1.2\columnwidth]{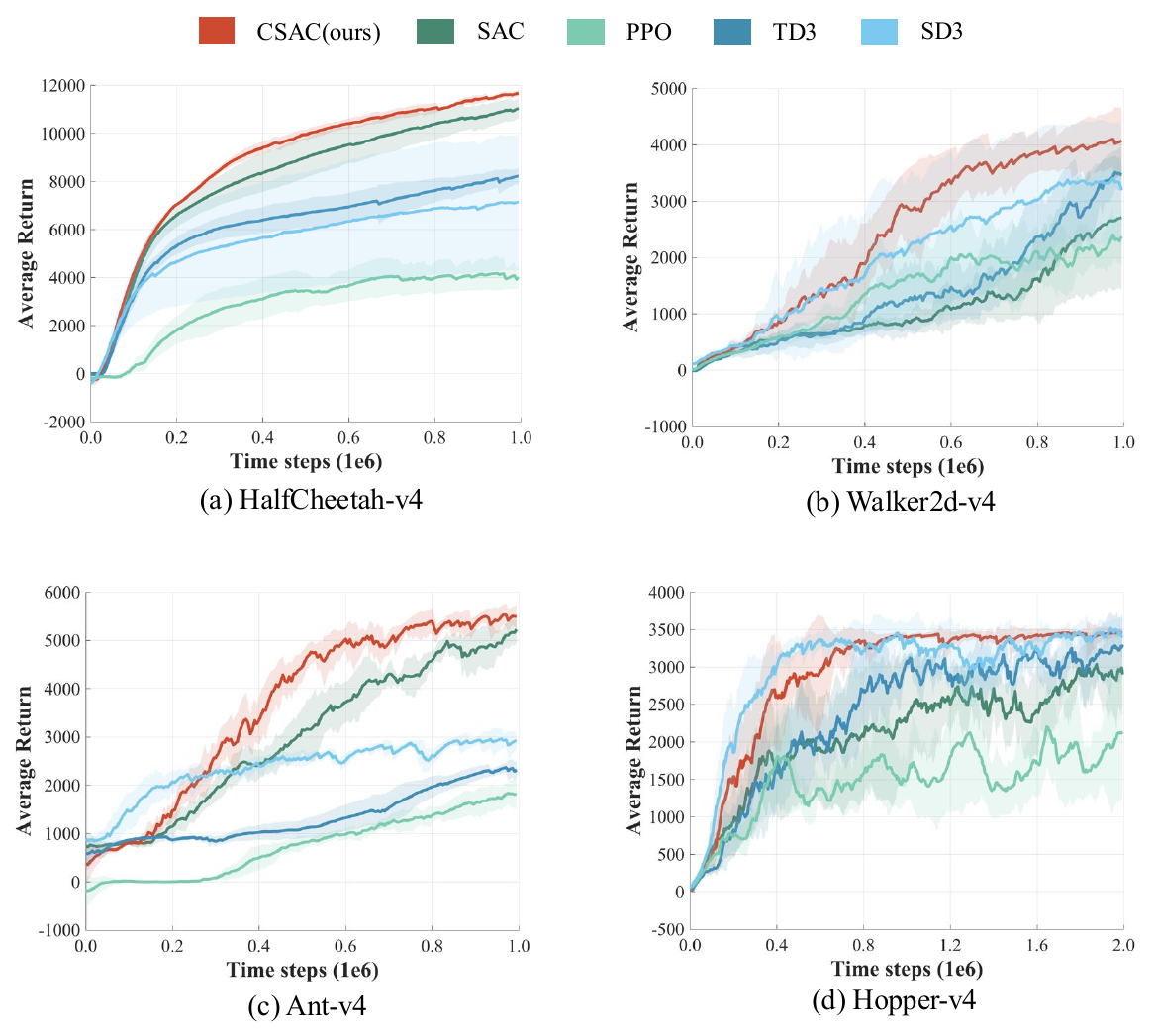}
	\caption{Learning curves over four benchmark tasks.  Curves are uniformly smoothed for visual clarity. The shaded region represents the corresponding standard deviation.}
	\label{lc mujoco}
\end{figure*}

The learning curves and maximum average returns are shown in Fig.~\ref{lc mujoco} and Table~\ref{table:performance} (best results in red). CSAC achieved the highest returns in HalfCheetah-v4, Walker2d-v4, and Ant-v4, outperforming SAC by 5.6\%, 51.5\%, and 5.9\%, respectively, and surpassing PPO, TD3, and SD3 by wider margins. In Walker2d-v4, where balancing and gait control are particularly challenging, CSAC's relative entropy regularization effectively prevented excessive policy updates, yielding the largest improvement over baselines. In Hopper-v4, CSAC performed comparably to SD3 (3458.22 vs.\ 3515.30) while still substantially outperforming SAC (+13.9\%) and PPO (+56.9\%). Overall, the joint entropy and relative entropy regularization enables CSAC to balance exploration and exploitation effectively, leading to superior and stable convergence across diverse continuous control tasks.

\begin{table*}[t]
    \caption{Maximum average returns over four benchmark tasks}
    \centering
    \label{table:performance}
    \resizebox{1.0\linewidth}{!}{\begin{tabular}{|c|c|c|c|c|c|}
        \hline
        & \textbf{CSAC (ours)} & \textbf{SAC} & \textbf{PPO} & \textbf{TD3} & \textbf{SD3} \\\hline
        HalfCheetah-v4 & \textbf{\textcolor{red}{11672.74 ± 151.21}} & 11052.89 ± 422.84 & 4181.89 ± 723.54 & 8224.70 ± 337.13 & 7158.19 ± 2821.74 \\\hline
        Walker2d-v4    & \textbf{\textcolor{red}{4106.21 ± 501.45}}  & 2710.65 ± 1272.01 & 2403.55 ± 502.86 & 3513.25 ± 295.81 & 3404.03 ± 989.50  \\\hline
        Ant-v4         & \textbf{\textcolor{red}{5538.20 ± 209.62}}  & 5229.39 ± 269.52  & 1836.47 ± 233.91 & 2368.35 ± 99.15  & 2960.61 ± 212.54  \\\hline
        Hopper-v4      & 3458.22 ± 59.87   & 3037.19 ± 242.06  & 2204.01 ± 540.71 & 3296.80 ± 356.86 & \textbf{\textcolor{red}{3515.30 ± 107.84}}  \\\hline
    \end{tabular}}
\end{table*}

\subsection{Ablation Study on Walker2d Task}\label{S4-3}

\begin{figure}[H]
	\centering
	\includegraphics[width=1.0\columnwidth]{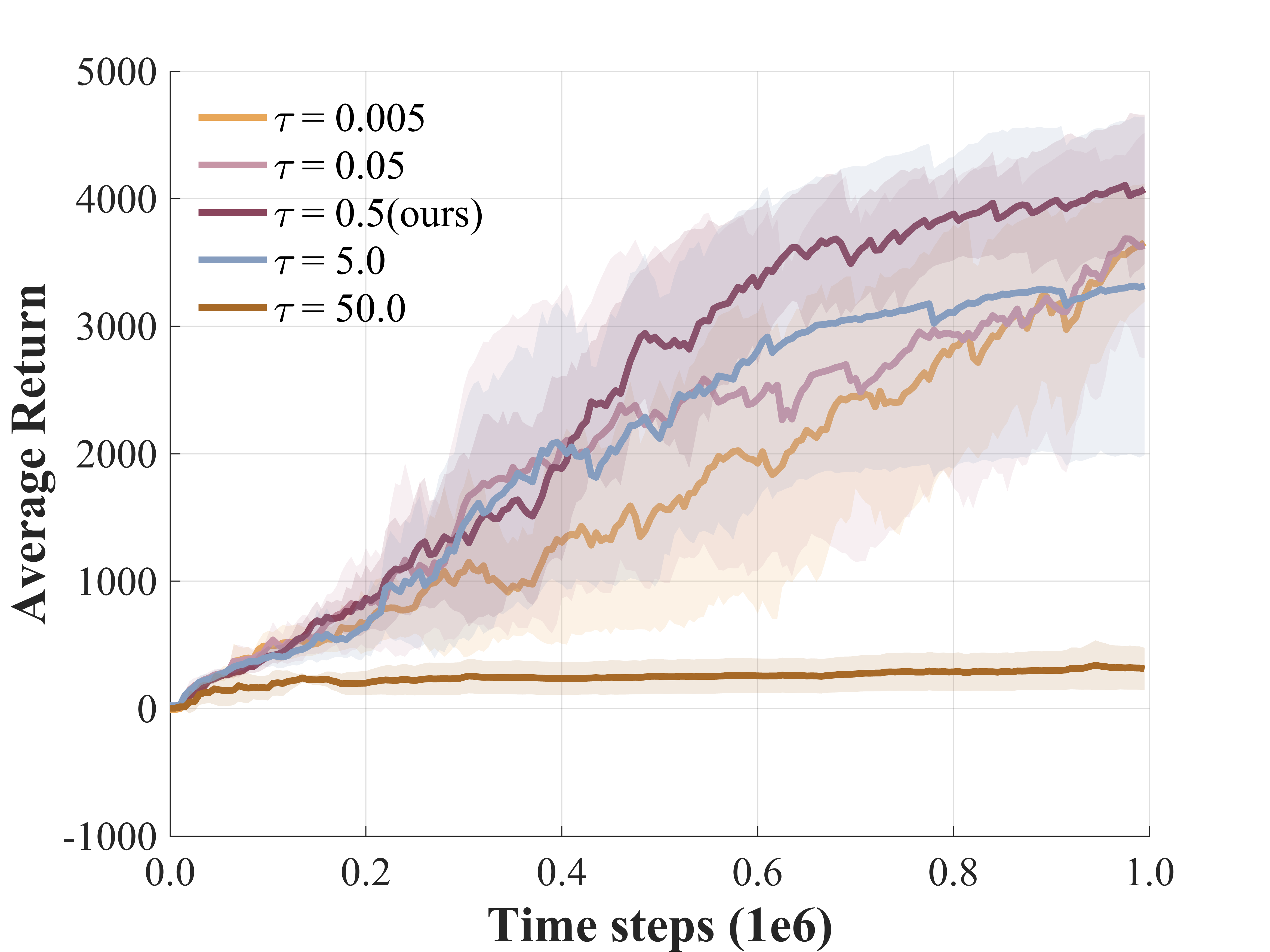}
	\caption{Learning curves of CSAC with different \(\tau\) values in the Walker2d-v4 task. The shaded region represents the standard deviation of the average evaluation over five trials. Curves are smoothed uniformly for visual clarity.}
	\label{fig:diff}
\end{figure}

Since the effect of the entropy coefficient has been extensively studied in SAC-related literature~\cite{haarnoja2018soft, haarnoja2018soft2}, we focus on the relative entropy coefficient $\tau$. Fig.~\ref{fig:diff} shows CSAC's learning curves in Walker2d-v4 under $\tau$ values ranging from 0.005 to 50.0.

The results reveal a clear sensitivity to $\tau$. Small values ($\tau = 0.005, 0.05$) weaken the reverse KL divergence constraint, causing the policy to focus on high-probability regions and reducing exploration, which leads to suboptimal convergence. The default setting $\tau = 0.5$ achieves the best performance, with final returns approaching 4000, effectively balancing exploration and exploitation. Conversely, large values ($\tau = 5.0, 50.0$) over-constrain policy updates, making the policy overly conservative and resulting in stagnating returns. These results confirm that a moderate $\tau$ is critical for optimal performance.

\subsection{Evaluation of Sample Efficiency}\label{S4-4}
\begin{figure}[H]
	\centering
	\includegraphics[width=1.0\columnwidth]{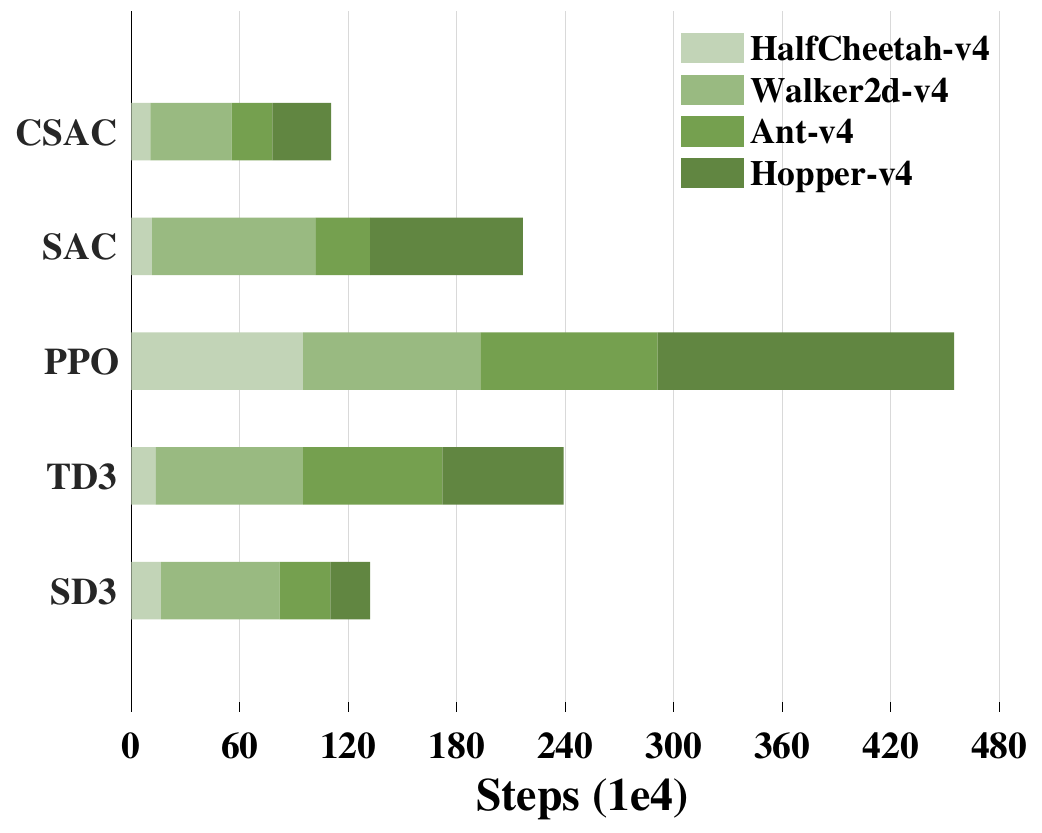}
	\caption{Number of interactions used by all compared approaches to reach the lower boundary of the maximum average returns over four benchmark control tasks.}
	\label{fig:step}
\end{figure}

We define sample efficiency as the number of interactions required to reach the lower boundary of the maximum average returns in Table~\ref{table:performance}---a metric critical for real-world deployment where sampling costs are high. Results are presented in Fig.~\ref{fig:step} and Table~\ref{table:episode_lengths} (best results in red).

\begin{table}[H]
    \caption{Number of Interactions Required to Reach the Lower Bound of Maximum Average Return (in $10^4$ Steps) }
    \centering
    \label{table:episode_lengths}
    \resizebox{1.0\linewidth}{!}{\begin{tabular}{|c|c|c|c|c|}
        \hline
        & \textbf{HalfCheetah-v4} & \textbf{Walker2d-v4} & \textbf{Ant-v4} & \textbf{Hopper-v4} \\\hline
        CSAC & \textbf{\textcolor{red}{10.5}} & \textbf{\textcolor{red}{45}} & \textbf{\textcolor{red}{22.5}} & 32.5 \\\hline
        SAC  & 11.5 & 90.5 & 30 & 84.5 \\\hline
        PPO  & 95   & 98   & 98 & 164  \\\hline
        TD3  & 13.5 & 81.5 & 77 & 67   \\\hline
        SD3  & 16.5 & 65.5 & 28 & \textbf{\textcolor{red}{22}}   \\\hline
    \end{tabular}}
\end{table}

CSAC achieved the fewest interaction steps in three of four tasks. In HalfCheetah-v4, it required only 105,000 steps, reducing the count by 8.7\% vs.\ SAC and over 8-fold vs.\ PPO. In Walker2d-v4 and Ant-v4, CSAC improved sample efficiency over SAC by 50.3\% and 25\%, respectively, and outperformed all other baselines by even wider margins. In Hopper-v4, SD3 was slightly more efficient (220,000 vs.\ 325,000 steps), yet CSAC still reduced steps by 61.5\% compared to SAC. These results confirm that the joint entropy and relative entropy regularization enables CSAC to converge rapidly with fewer environment interactions across diverse tasks.

\subsection{Evaluation of Robustness in HalfCheetah Task}\label{S4-5}

\begin{figure*}[t]
	\centering
	\includegraphics[width=1.8\columnwidth]{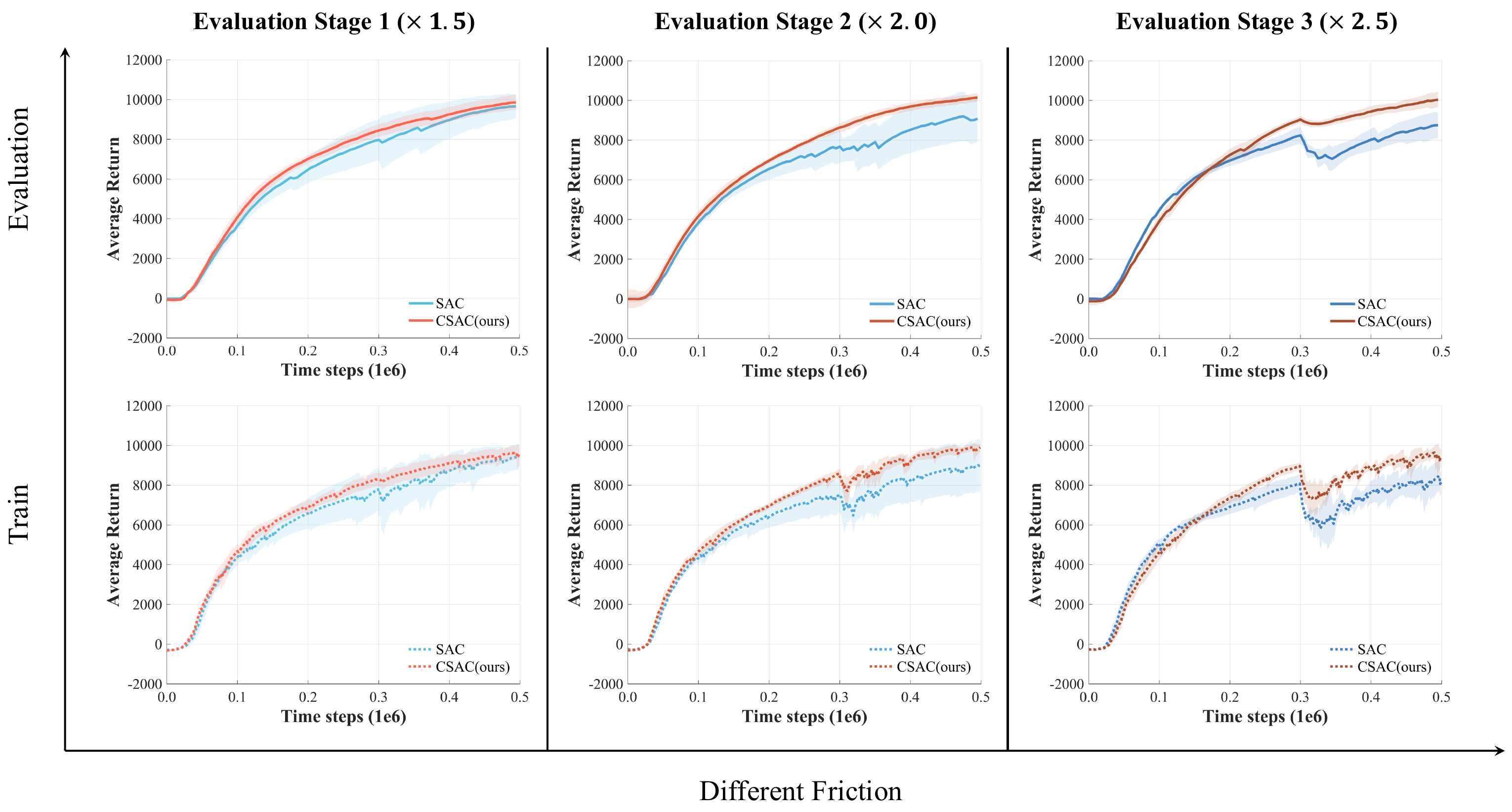}
	\caption{Performance comparison of CSAC and SAC under varying Friction-Induced dynamic noise in the HalfCheetah-v4 environment. The shaded region represents the corresponding standard deviation.}
	\label{fig:stable}
\end{figure*}

To assess robustness under dynamic perturbations, we compared CSAC with SAC---chosen for its structural similarity---in HalfCheetah-v4, where friction was suddenly increased to 1.5$\times$, 2.0$\times$, and 2.5$\times$ its original value after 300,000 steps of standard training. The results are shown in Fig.~\ref{fig:stable}.

At 1.5$\times$ friction (left panel), both algorithms experienced return fluctuations, but CSAC recovered faster and maintained higher evaluation returns. As friction rose to 2.0$\times$ (middle panel), SAC's performance became increasingly erratic, whereas CSAC recovered more quickly and consistently sustained a higher return level during evaluation. At the most challenging 2.5$\times$ friction (right panel), CSAC again exhibited a smaller performance decline and faster adaptation, while SAC showed severe instability and failed to recover to its initial level. Across all noise intensities, CSAC consistently demonstrated faster recovery and greater stability than SAC, confirming the effectiveness of relative entropy regularization in enhancing robustness under dynamic perturbations.

\subsection{Implementation on Real-Robot-Based Simulation}\label{S4-6}

\subsubsection{Experimental Settings for Real-Robot-Based Simulation Experiments}\label{S4-6-1}

\begin{figure}[t]
	\centering
	\includegraphics[width=1.0\columnwidth]{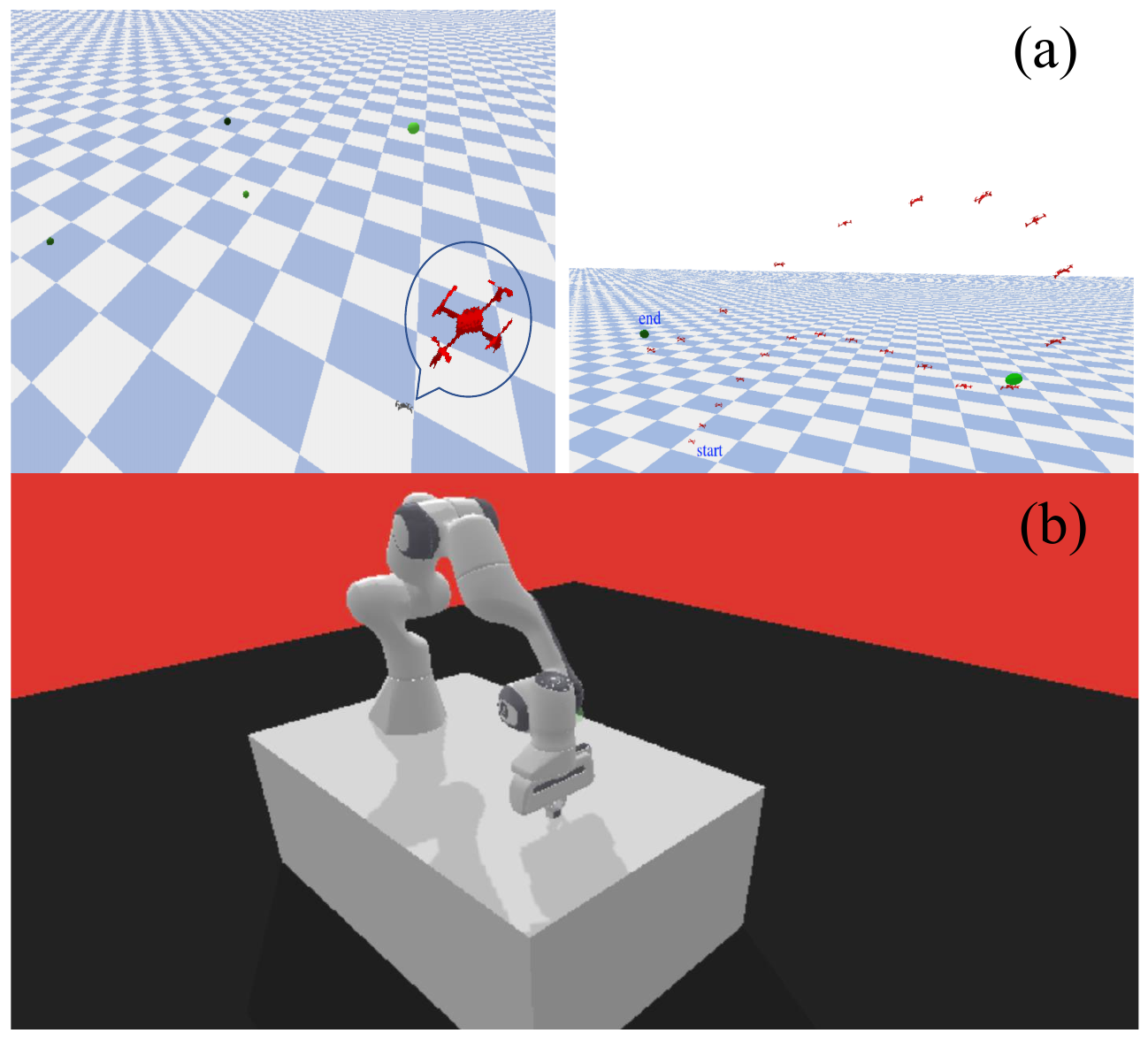}
	\caption{Real-Robot-Based simulation experiments for evaluation in this article. (a) QuadX-Waypoints-v1. (b) PandaReach-v2.}
	\label{fig:robot}
\end{figure}

We employed two PyBullet-based simulation platforms~\cite{coumans2021}---PyFlyt~\cite{tai2023pyflyt} and Panda Gym~\cite{gallouedec2021pandagym}---to evaluate CSAC in real-robot-based simulations (Fig.~\ref{fig:robot}, Table~\ref{table:robot ob}). The QuadX-Waypoints-v1 task (Fig.~\ref{fig:robot}a) requires a quadrotor UAV to navigate through four designated waypoints, while the PandaReach-v2 task (Fig.~\ref{fig:robot}b) requires a Panda robotic arm to move its end-effector to a target position.

\begin{table}[t]
    \caption{Observation and Action Spaces for Two Real-Robot-Based Simulation Tasks}
    \centering
    \resizebox{1.0\linewidth}{!}{\begin{tabular}{|c|c|c|}
        \hline
        \textbf{} & \textbf{Observation Space} & \textbf{Action Space} \\\hline
        QuadX-Waypoints-v1 & 9  & 4 \\\hline
        PandaReach-v2      & 17 & 6 \\\hline
    \end{tabular}}
     \label{table:robot ob}
\end{table}

Common parameters follow Table~\ref{table:parameters}; CSAC-specific hyperparameters ($\sigma$, $\tau$) are listed in Table~\ref{table:spe para}. All experiments were averaged over five independent trials.

\begin{table}[H]
    \caption{SPECIFIC HYPERPARAMETERS OF CSAC}
    \centering
    \resizebox{0.6\linewidth}{!}{\begin{tabular}{|c|c|c|}
        \hline
        \textbf{} & \boldmath$\sigma$ & \boldmath$\tau$ \\\hline
        QuadX-Waypoints-v1 & 0.2 & 0.1 \\\hline
        PandaReach-v2      & 0.2 & 0.5 \\\hline
    \end{tabular}}
    \label{table:spe para}
\end{table}

\subsubsection{Evaluation of the Learning Capability on Two Real-Robot-Based Simulation Tasks}\label{S4-6-2}

\begin{figure*}[t]
	\centering
	\includegraphics[width=1.5\columnwidth]{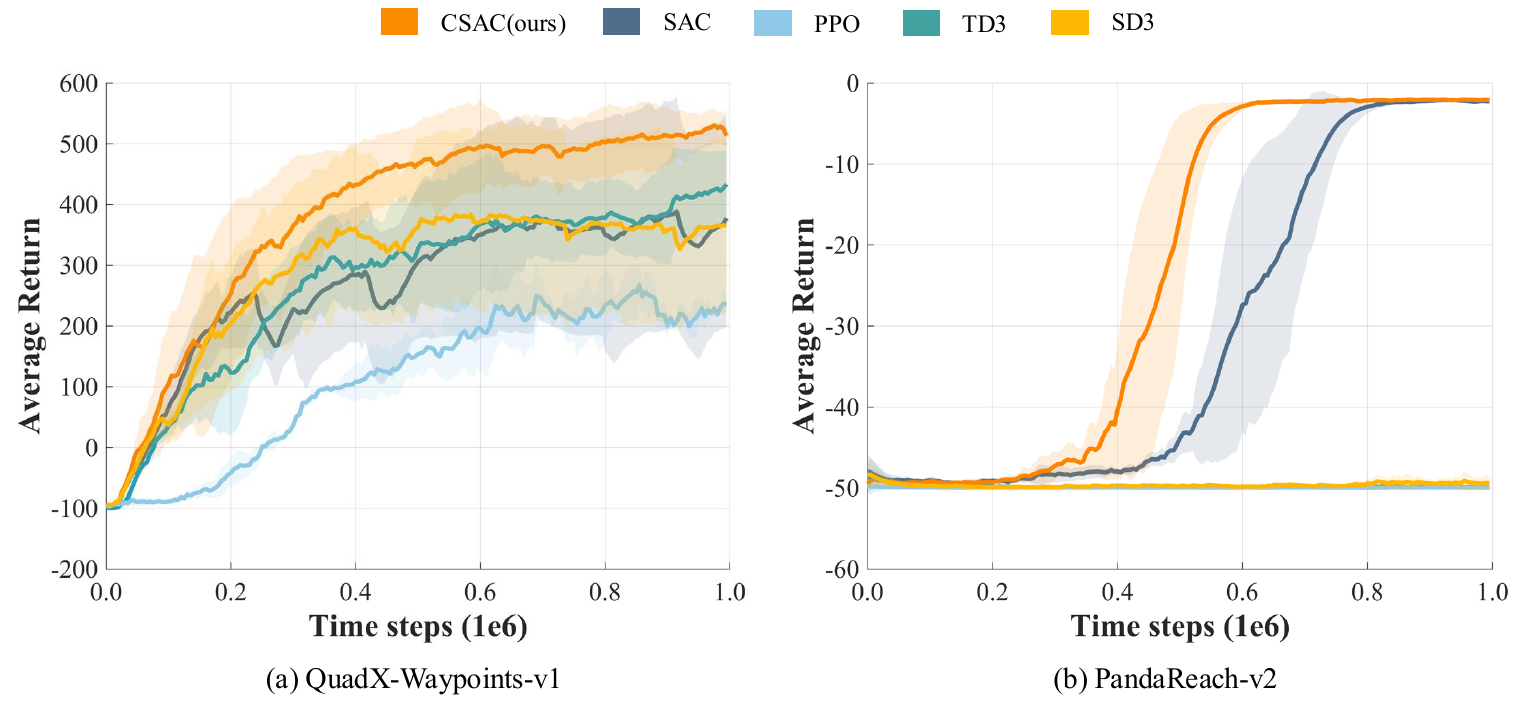}
	\caption{Learning curves of two Real-Robot-Based simulation tasks. The shaded region represents the standard deviation of the average evaluation over five independent trials.Curves are uniformly smoothed for visual clarity.}
	\label{fig:robot lc}
\end{figure*}

The learning curves and maximum average returns are shown in Fig.~\ref{fig:robot lc} and Table~\ref{table:robot lc} (best results in red).

\begin{table*}[t]
    \caption{MAX AVERAGE RETURNS OVER two Real-Robot-Based simulation TASKS}
    \centering
    \resizebox{1.0\linewidth}{!}{\begin{tabular}{|c|c|c|c|c|c|}
        \hline
        & \textbf{CSAC (ours)} & \textbf{SAC} & \textbf{PPO} & \textbf{TD3} & \textbf{SD3} \\\hline
        QuadX-Waypoints-v1 & \textbf{\textcolor{red}{530.28 ± 29.04}} & 388.55 ± 189.51 & 267.06 ± 32.07 & 433.56 ± 56.03 & 383.03 ± 145.51 \\\hline
        PandaReach-v2      & \textbf{\textcolor{red}{-2.04 ± 0.18}}   & -2.12 ± 0.22   & -49.80 ± 0.04  & -48.57 ± 0.08  & -48.36 ± 0.10  \\\hline
    \end{tabular}}
    \label{table:robot lc}
\end{table*}

In QuadX-Waypoints-v1, CSAC achieved a final average return of 530.28, outperforming SAC by 36.5\%, TD3 by 22.3\%, PPO by 98.6\%, and SD3 by 38.5\%. In PandaReach-v2, which demands high-precision control, CSAC ($-2.04$) slightly improved upon SAC ($-2.12$) while exhibiting lower variance; PPO, TD3, and SD3 all performed poorly (returns $< -48$), highlighting their deficiency in fine control tasks. Overall, CSAC demonstrated superior and stable performance across both real-robot-based tasks.

\subsubsection{Evaluation of Robustness in Drones Control}\label{S4-6-3}

\begin{figure}[H]
	\centering
	\includegraphics[width=1.0\columnwidth]{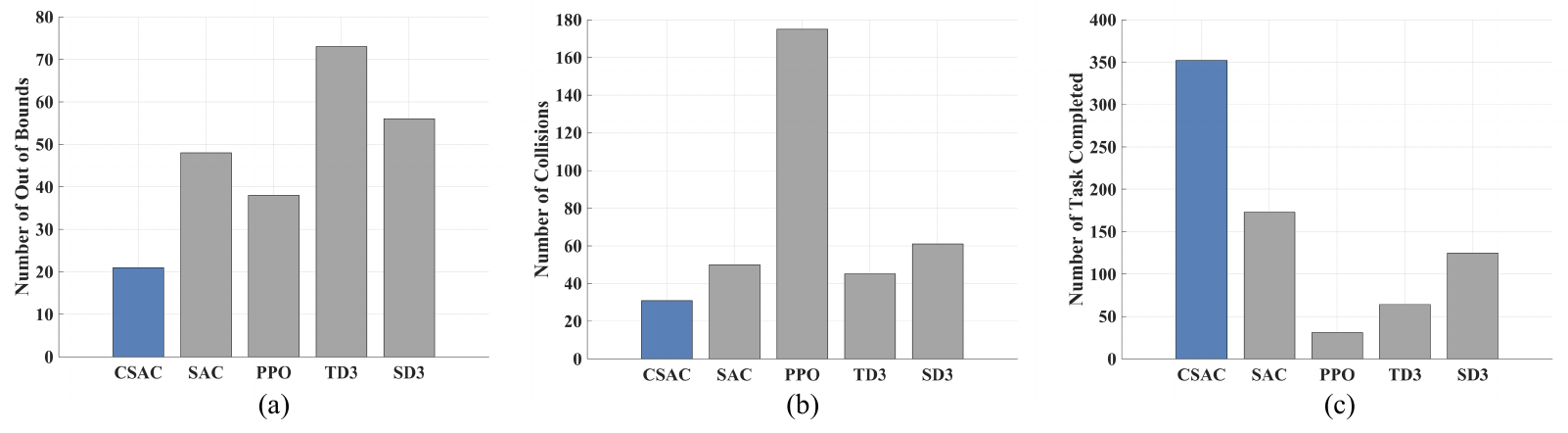}
	\caption{The average number of each condition in QuadX-Waypoints-v1: (a) Number of out of bounds; (b) Number of collisions; (c) Number of task completed.}
	\label{ra}
\end{figure}

\begin{table}[H]
    \caption{Case Analysis in QuadX-Waypoints-v1}
    \centering
    \label{table:robust}
    \resizebox{1.0\linewidth}{!}{\begin{tabular}{|c|c|c|c|}
        \hline
        & \textbf{Out of Bound} & \textbf{Collision} & \textbf{Completion} \\\hline
        CSAC & \textbf{\textcolor{red}{20.6 ± 11.24}} & \textbf{\textcolor{red}{31.6 ± 10.41}} & \textbf{\textcolor{red}{352.0 ± 241.57}} \\\hline
        SAC  & 48.2 ± 27.54 & 50.6 ± 14.36 & 173.4 ± 230.56 \\\hline
        PPO  & 38.4 ± 17.93 & 175.4 ± 4.16 & 31.0 ± 19.67 \\\hline
        TD3  & 73.2 ± 20.5  & 45.4 ± 15.01 & 64.0 ± 68.31 \\\hline
        SD3  & 56.0 ± 18.72 & 61.2 ± 24.34 & 125.4 ± 158.86 \\\hline
    \end{tabular}}
\end{table}

We evaluated the robustness of learned policies in QuadX-Waypoints-v1 by recording out-of-bounds, collision, and successful completion counts over 1000 evaluation rollouts per trial (five trials total). Results are shown in Fig.~\ref{ra} and Table~\ref{table:robust} (best results in red). CSAC achieved the fewest failures and the most completions among all methods, reducing out-of-bounds by 46.3\% vs.\ PPO and collisions by 30.4\% vs.\ TD3, while completing over 350 rollouts successfully---close to the sum of all four baselines combined. These results confirm that the joint regularization enables effective exploration while preventing the overly aggressive policy updates that lead to unstable drone control.

\subsubsection{Evaluation of Effectiveness in Robotic Arm}\label{S4-6-4}

\begin{figure}[t]
	\centering
	\includegraphics[width=1.0\columnwidth]{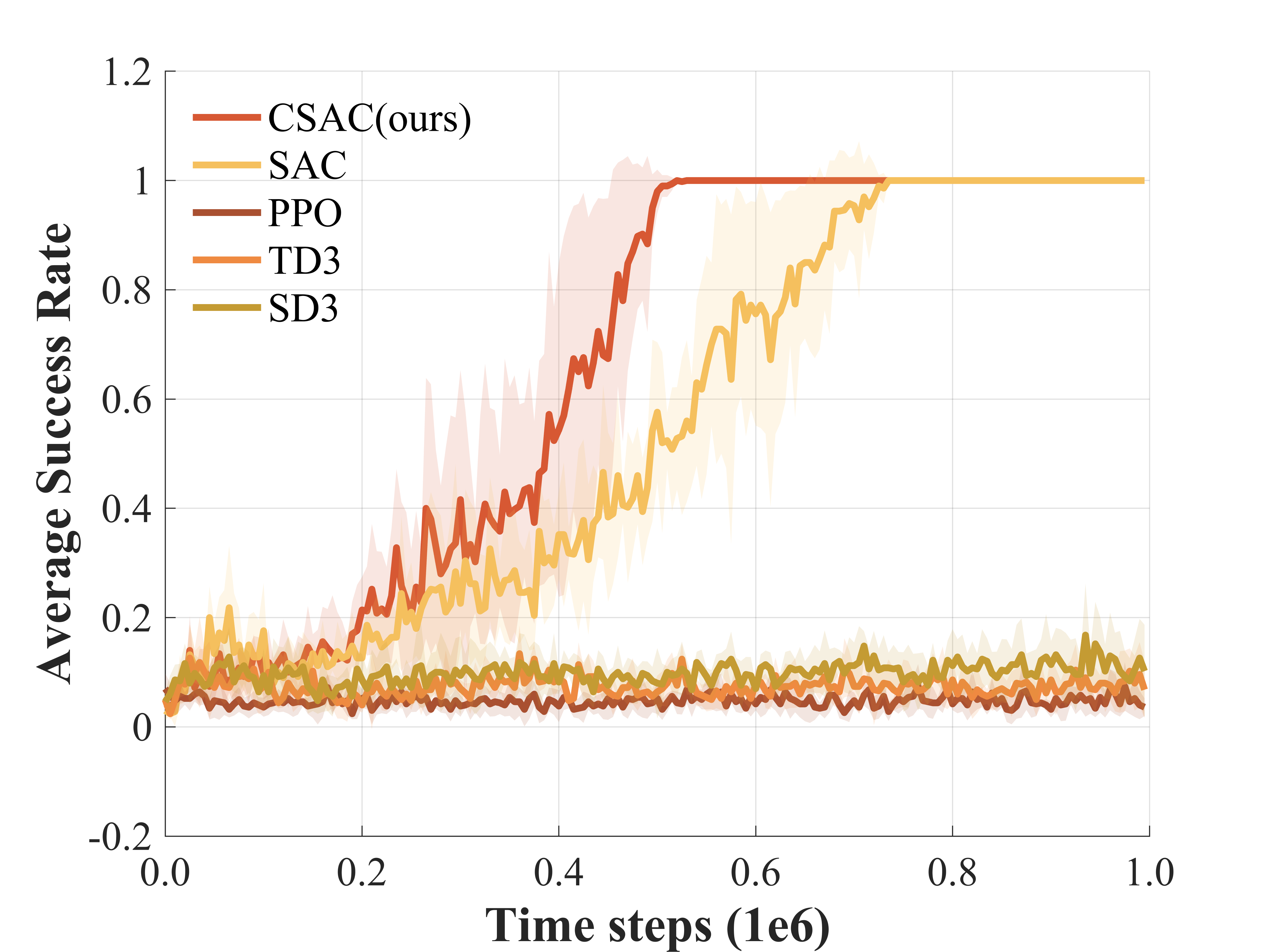}
	\caption{The average successs rate in PandaReach-v2.}
	\label{fig:panda_success}
\end{figure}

We further evaluated the average success rate in PandaReach-v2 (Fig.~\ref{fig:panda_success}). CSAC's success rate rose rapidly around 400,000 steps and reached nearly 100\% by 600,000 steps, whereas SAC converged noticeably slower (around 800,000 steps) with a slightly lower final rate. PPO, TD3, and SD3 exhibited large fluctuations and failed to achieve effective policy learning, indicating their inability to meet the high-precision demands of this task. These results confirm CSAC's effectiveness and fast convergence in fine control scenarios.

\section{Conclusion}\label{S5}
This paper introduces Conservative Soft Actor-Critic (CSAC), which integrates entropy and relative entropy regularization into the Actor-Critic framework to enhance both exploration and stability for robot control. Experiments on four MuJoCo benchmark tasks and two real-robot-based simulation tasks demonstrate that CSAC achieves superior learning capability, sample efficiency, and robustness over state-of-the-art baselines, confirming the benefit of jointly leveraging both regularization terms.

For future work, we plan to develop adaptive strategies for dynamically adjusting $\sigma$ and $\tau$ to optimize the exploration-exploitation trade-off, and to deploy CSAC on real-world hardware such as UAVs and robotic arms to validate its effectiveness under physical constraints and real-time decision-making requirements.






\bibliographystyle{IEEEtran}
\bibliography{paper}

\end{document}